\documentclass{article}
\usepackage{booktabs} 
\usepackage{authblk}
\usepackage{amsmath}
\usepackage{amsfonts}
\usepackage{amssymb}
\usepackage{graphicx}
\usepackage{hyperref}
\usepackage{float}
\usepackage{array} 
\usepackage{multirow}
\usepackage{listings}
\usepackage[top=3cm, bottom=3cm, left=2.5cm, right=2.5cm]{geometry} 
\usepackage[round, authoryear]{natbib} 
\usepackage{url} 

\newcolumntype{M}[1]{>{\centering\arraybackslash}m{#1}}

\begin{document}

\title{Think Twice: Enhancing LLM Reasoning by Scaling Multi-round Test-time Thinking}
\author{Xiaoyu Tian}
\author{Sitong Zhao}
\author{Haotian Wang}
\author{Shuaiting Chen}
\author{Yunjie Ji}
\author{Yiping Peng}
\author{Han Zhao}
\author{Xiangang Li}

\affil{
    \raisebox{-0.4em}{\includegraphics[height=1.5em]{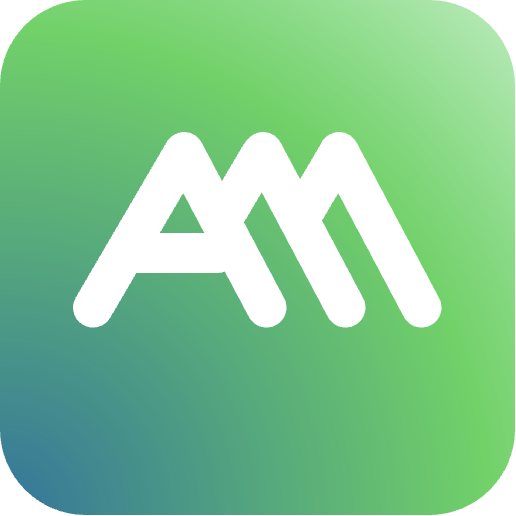}}
    \hspace{0.2em}a-m-team
}
\date{}

\maketitle

\begin{abstract}


Recent advances in large language models (LLMs), such as OpenAI-o1 and DeepSeek-R1, have demonstrated the effectiveness of test-time scaling, where extended reasoning processes substantially enhance model performance. Despite this, current models are constrained by limitations in handling long texts and reinforcement learning (RL) training efficiency. To address these issues, we propose a simple yet effective test-time scaling approach—\textbf{Multi-round Thinking}. This method iteratively refines model reasoning by leveraging previous answers as prompts for subsequent rounds. Extensive experiments across multiple models, including QwQ-32B and DeepSeek-R1, consistently show performance improvements on various benchmarks such as AIME 2024, MATH-500, GPQA-diamond, and LiveCodeBench. For instance, the accuracy of QwQ-32B improved from 80.3\% (Round 1) to 82.1\% (Round 2) on the AIME 2024 dataset, while DeepSeek-R1 showed a similar increase from 79.7\% to 82.0\%. These results confirm that \textbf{Multi-round Thinking} is a broadly applicable, straightforward approach to achieving stable enhancements in model performance, underscoring its potential for future developments in test-time scaling techniques.

The key prompt:
\begin{quote}
\ttfamily
\textit{Original question prompt}\\
The assistant's previous answer is: <answer> \textit{last round answer} </answer>, and please re-answer.
\end{quote}

\begin{figure}[h!]
    \centering
    \includegraphics[width=0.75\linewidth]
    {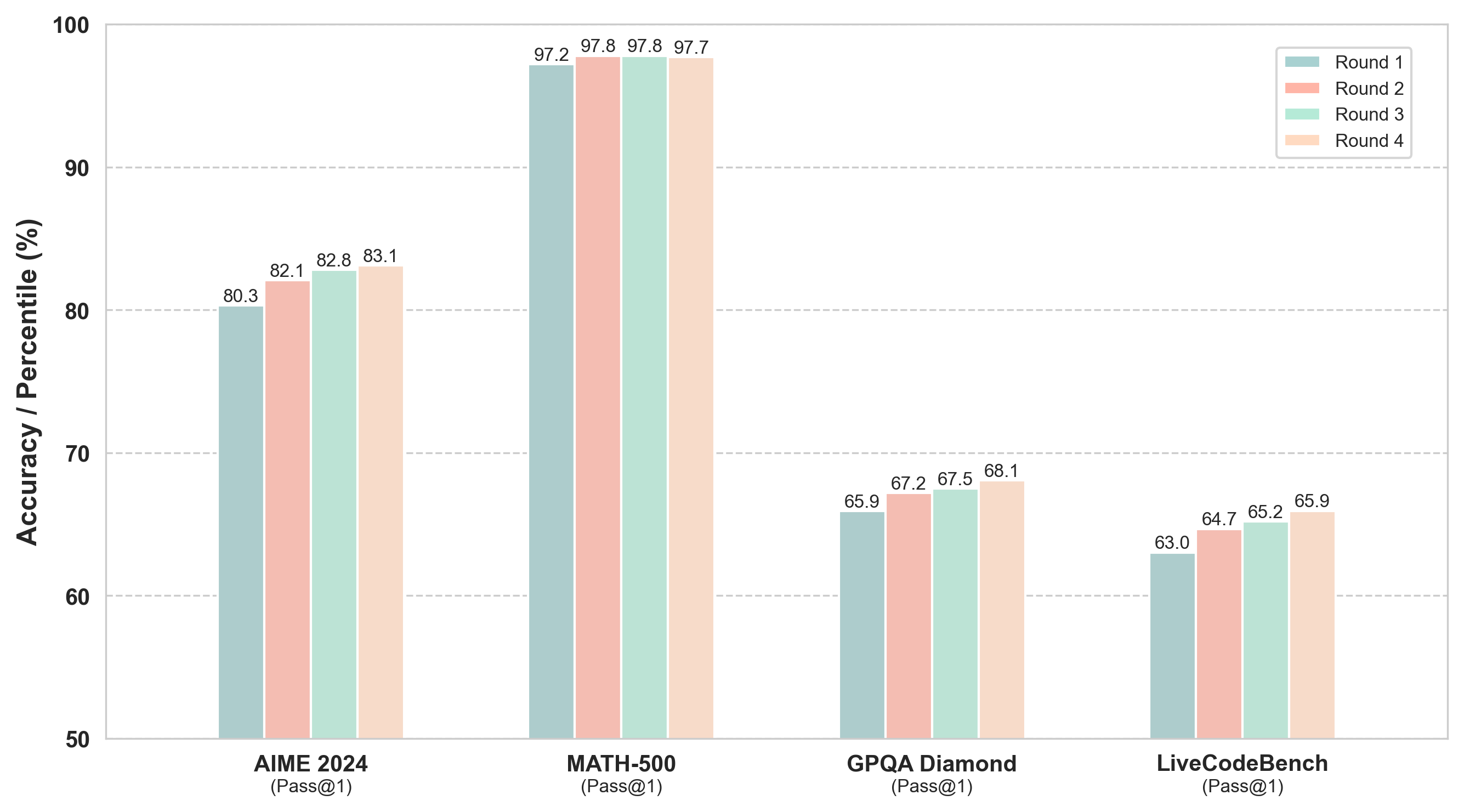}
    \caption{Benchmark performance of QwQ-32B using Multi-round Thinking.}
    \label{fig:QwQ-32B res}
\end{figure}

\begin{figure}[h!]
    \centering
    \includegraphics[width=0.75\linewidth]{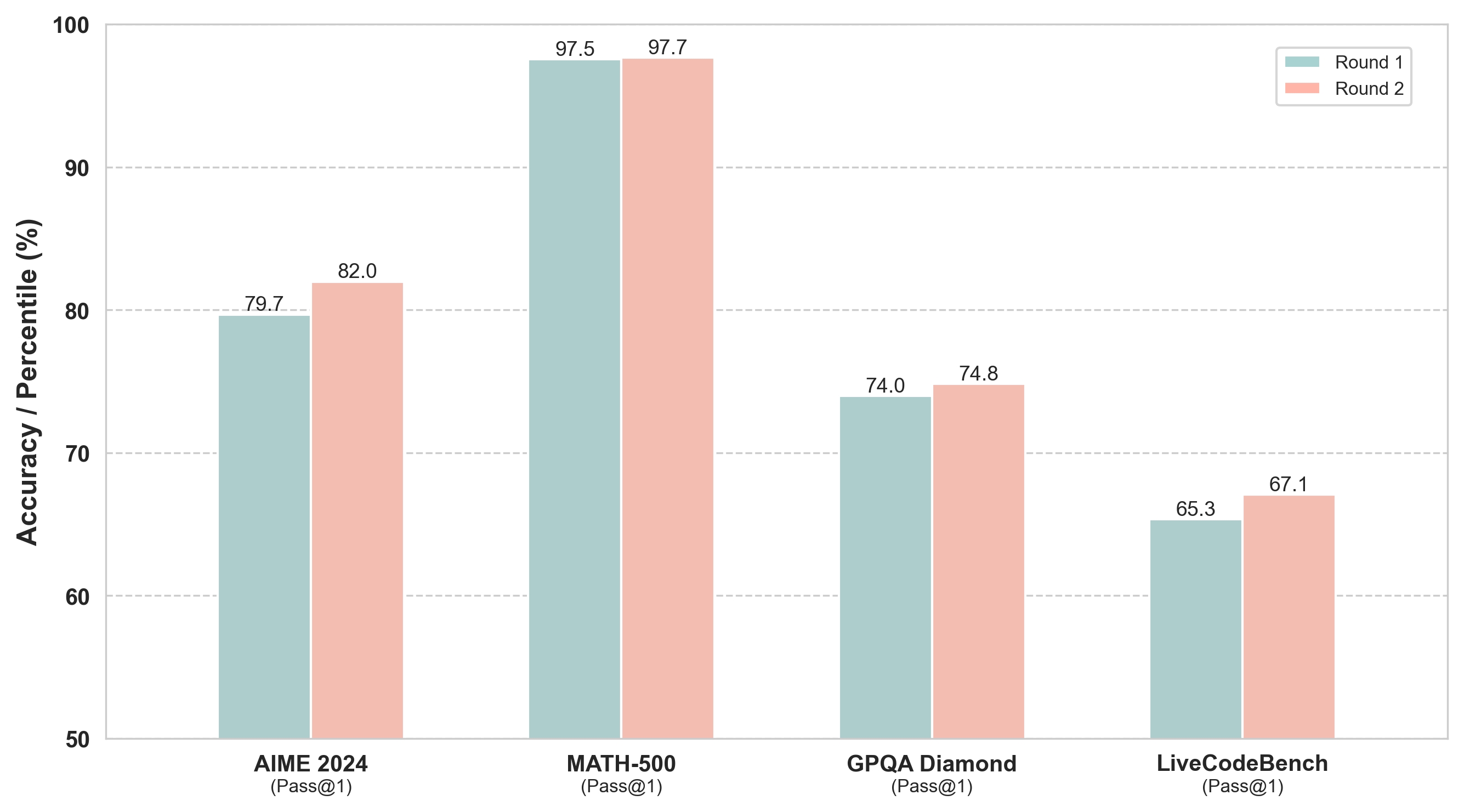}
    \caption{Benchmark performance of DeepSeek-R1 using Multi-round Thinking.}
    \label{fig:DS res}
\end{figure}
\end{abstract}

\section{Introduction}

Inference test-time compute \citep{yang2025thinkingoptimalscalingtesttimecompute, wu2025inferencescalinglawsempirical} refers to the computational resources utilized by large language models (LLMs) during the generation of prompt responses, distinct from the training compute used for model creation and refinement. Leveraging step-by-step reasoning has shown substantial improvements in solving complex tasks by explicitly providing models with intermediate reasoning steps\citep{lightman2023letsverifystepstep, wei2023chainofthoughtpromptingelicitsreasoning}, significantly enhancing accuracy.

In recent years, the performance improvements of language models have largely depended on massive-scale self-supervised pre-training \citep{kaplan2020scalinglawsneurallanguage, hoffmann2022trainingcomputeoptimallargelanguage}, scaling up training-time compute. However, as advancements in training-time scaling slow, increasing attention is turning towards scaling up test-time compute \citep{muennighoff2025s1simpletesttimescaling,chen2025reasoningerasurveylong}. OpenAI \citep{OpenAI2024} pioneered this approach with their o1 series models \citep{openai2024openaio1card} using large-scale reinforcement learning (RL). 

DeepSeek further advanced test-time scaling by introducing the DeepSeek-R1 \citep{deepseekai2025deepseekr1incentivizingreasoningcapability}, successfully achieving performance comparable to OpenAI's o1 series. Prior approaches in inference test-time compute have included majority voting methods and external reward-based best-of-N strategies \citep{levi2024simplemodelinferencescaling, diao2024activepromptingchainofthoughtlarge}. Unlike repetitive sampling, sequential expansion approaches enable models to iteratively refine attempts based on prior outcomes. Many researchers have attempted to replicate or extend their methods, employing Monte Carlo Tree Search (MCTS) \citep{zhou2024languageagenttreesearch, choi2023kctsknowledgeconstrainedtreesearch}, multi-agent approaches \citep{qin2024o1replicationjourneystrategic, li2025searcho1agenticsearchenhancedlarge}, some work based on Process Reward Model(PRM) \citep{wang2024mathshepherdverifyreinforcellms, lightman2023letsverifystepstep}.

Despite these successes, existing methods exhibit critical limitations. PRM face challenges such as defining fine-grained reasoning steps clearly, verifying intermediate reasoning correctness, and mitigating reward hacking \citep{amodei2016concreteproblemsaisafety, langosco2023goalmisgeneralizationdeepreinforcement}, making automated labeling challenging and manual labeling impractical for scaling. Similarly, MCTS methods encounter difficulties due to vast search spaces, often causing models to become trapped in local optima, and depend heavily on sophisticated scoring models that are challenging to train \citep{deepseekai2025deepseekr1incentivizingreasoningcapability}. 

Addressing these issues, DeepSeek introduced a rule-based reward system combined with large-scale reinforcement learning (RL), enabling clearer guidance and promoting model self-reflection and deeper reasoning \citep{deepseekai2025deepseekr1incentivizingreasoningcapability}. However, consistently identifying optimal reasoning paths remains challenging.

Inspired by human cognitive behaviors, we propose a novel test-time scaling strategy named \textbf{Multi-round Thinking}. This method allows the model to iteratively reconsider previous answers independently, using only the final answer from previous rounds as input prompts, discarding prior reasoning steps. This approach parallels human cognitive processes, breaking cognitive inertia and enabling the model to correct entrenched reasoning errors.

Our experimental results demonstrate the effectiveness of this intuitive approach. For example, using the DeepSeek-R1 model \citep{deepseekai2025deepseekr1incentivizingreasoningcapability}, performance improvements were observed across multiple benchmarks: on AIME 2024 \citep{maa_aime_2024}, pass@1 increased from 79.7\% (Round 1) to 82.0\% (Round 2); on GPQA-Diamond \citep{rein2023gpqagraduatelevelgoogleproofqa}, it rose from 74.0\% to 74.8\%; and on LiveCodeBench \citep{jain2024livecodebench}, performance improved from 65.3\% to 67.1\%. These findings underscore the substantial potential of iterative thinking for further exploiting the benefits of test-time scaling.

\section{Approach}

We introduce a novel \textbf{Multi-round Thinking} approach designed to significantly enhance reasoning capabilities in large language models (LLMs). In contrast to traditional single-step reasoning methods, our approach iteratively refines answers through multiple rounds of inference. Each round takes the answer from the previous iteration (without intermediate reasoning steps) as part of a new input prompt, encouraging independent reconsideration and correction. This iterative process helps models avoid cognitive inertia, analogous to human strategies in overcoming entrenched errors in reasoning.

The Multi-round Thinking methodology operates explicitly as follows:

Given an original user prompt \(P_{user}\), the inference and refinement process proceeds iteratively:

\textbf{Initial Round (Round 1):}  
The language model receives the initial prompt and generates the first round of reasoning and final answer:
\begin{equation}
M(P_{user}) \rightarrow \{Thinking_{1}, Answer_{1}\}
\end{equation}

\textbf{Subsequent Rounds (Round \(n\), \(n \geq 2\)):}  
In each subsequent inference round, intermediate reasoning traces (\(Thinking_{n-1}\)) from the previous iteration are discarded, retaining only the final answer (\(Answer_{n-1}\)). The prompt for the next round is constructed by concatenating the original user prompt and the previously obtained answer:
\begin{equation}
P_{n} = P_{user} \oplus Answer_{n-1}
\end{equation}

The model independently reevaluates the newly formed prompt and produces an updated reasoning trace and refined answer:
\begin{equation}
M(P_{n}) \rightarrow \{Thinking_{n}, Answer_{n}\}
\end{equation}

This iterative refinement cycle can be formally represented as follows:
\begin{align}
&P_{1} = P_{user}, \quad M(P_{1}) \rightarrow \{Thinking_{1}, Answer_{1}\} \\
&P_{2} = P_{user} \oplus Answer_{1}, \quad M(P_{2}) \rightarrow \{Thinking_{2}, Answer_{2}\} \\
&P_{3} = P_{user} \oplus Answer_{2}, \quad M(P_{3}) \rightarrow \{Thinking_{3}, Answer_{3}\} \\
&\quad\quad\quad\quad\quad\quad\quad\quad\quad\quad\quad\quad \vdots \\
&P_{n} = P_{user} \oplus Answer_{n-1}, \quad M(P_{n}) \rightarrow \{Thinking_{n}, Answer_{n}\}
\end{align}

In these equations, \(\oplus\) denotes the textual concatenation operation used to form the iterative prompts. Through this repeated refinement procedure, the model is encouraged to reconsider previous conclusions independently, effectively minimizing cognitive inertia and systematically improving the quality of reasoning outcomes.

Specifically, for a given question prompt $P$, we first use a reasoning model to answer the question, producing a thought process $T$ and an answer $A$. Then, we concatenate the original question prompt $P$ and the answer $A$ using the following prompt:

\begin{quote}
\ttfamily
\textit{Original question prompt}\\
The assistant's previous answer is: <answer> \textit{last round answer} </answer>, and please re-answer.
\end{quote}

We then send this prompt to the large model again to generate a new answer. Using this method, we can obtain multi-turn responses.

\section{Experiments}
\subsection{Evaluation}
\subsubsection{Benchmark}
We evaluated the reasoning ability of the model using LiveCodeBench \citep{jain2024livecodebench}, GPQA-Diamond \citep{rein2023gpqagraduatelevelgoogleproofqa}, AIME 2024 \citep{maa_aime_2024}, and MATH-500 \citep{lightman2023letsverifystepstep}. These benchmarks span multiple fields and difficulty levels, enabling a thorough assessment of the model's reasoning performance across diverse scenarios.

\subsubsection{Evaluation Methodology}
We standardized the evaluation conditions by setting the maximum generation length at 32,768 tokens. For benchmarks requiring stochastic sampling, we uniformly set the temperature to 0.6 and the top-p value to 0.95. Specifically, for AIME 2024 \citep{maa_aime_2024}, we generated 32 samples per query to calculate pass@1 accuracy. For LiveCodeBench \citep{jain2024livecodebench} and GPQA-Diamond \citep{rein2023gpqagraduatelevelgoogleproofqa}, we generated 8 samples per query to estimate pass@1. For MATH-500 \citep{lightman2023letsverifystepstep}, we generated 4 responses per query, also to estimate pass@1 accuracy. The primary evaluation metric adopted was the global average accuracy across all benchmarks.

\subsection{Results and Analysis}

\subsubsection{Overall Results of Multi-round Thinking}
Experimental results comparing initial (Round 1) and Multi-round Thinking (Round 2) performance are summarized in Table 1.
\begin{table}[htbp]
  \caption{Model Performance Comparison (pass@1 accuracy) Between Single-round (Round 1) and Multi-round Thinking (Round 2-4) Across Different Benchmarks}
  \vspace{0.5em}
  \centering
  \renewcommand{\arraystretch}{1.2}
  \label{tab:model_performance}
  \resizebox{\textwidth}{!}{
    \begin{tabular}{l|>{\centering\arraybackslash}m{\dimexpr0.07\textwidth\relax}||
                      M{\dimexpr0.12\textwidth\relax}
                      M{\dimexpr0.12\textwidth\relax}
                      M{\dimexpr0.12\textwidth\relax}
                      M{\dimexpr0.18\textwidth\relax}|
                      M{\dimexpr0.12\textwidth\relax}}
      \hline
      \multicolumn{1}{l|}{\textbf{Model}} & \textbf{Round} & \textbf{AIME 2024 pass@1} & \textbf{MATH500 pass@1} & \textbf{GPQA-Diamond pass@1} & \textbf{LiveCodeBench pass@1} & \textbf{Average}\\[1.5ex]
      \hline
      \multirow{2}{*}{\textbf{Deepseek-R1}} 
        & 1 & 79.7 & 97.6 & 74.0 & 65.3 & 79.2\\
        & \textbf{2} & \textbf{82.0} & \textbf{97.6} & \textbf{74.8} & \textbf{67.1} & \textbf{80.4}\\
      \hline
      \multirow{2}{*}{\textbf{QwQ-32B}} 
        & 1 & 80.3 & 97.2 & 65.9 & 63.0 & 76.6\\
        & 2 & 82.1 & 97.8 & 67.2 & 64.7 & 78.0\\
        & 3 & 82.8 & 97.8 & 67.5 & 65.2 & 78.3\\
        & \textbf{4} & \textbf{83.1} & \textbf{97.7} & \textbf{68.1} & \textbf{66.0} & \textbf{78.7}\\
      \hline
      \multirow{2}{*}{\textbf{DeepSeek-R1-Distill-Qwen-32B}} 
        & 1 & 72.0 & 96.0 & 60.1 & 57.0 & 71.3\\
        & \textbf{2} & \textbf{75.1} & \textbf{96.3} & \textbf{61.3} & \textbf{57.6} & \textbf{72.6}\\
      \hline
      \multirow{2}{*}{\textbf{DeepSeek-R1-Distill-Qwen-7B}} 
        & 1 & 56.9 & 93.4 & 49.2 & 35.0 & 58.6\\
        & \textbf{2} & \textbf{58.4} & \textbf{93.9} & \textbf{49.4} & \textbf{36.7} & \textbf{59.6}\\
      \hline
      \multirow{2}{*}{\textbf{AM-Distill-Qwen-32B}} 
        & 1 & 72.8 & 96.2 & 62.3 & 58.3 & 72.4\\
        & \textbf{2} & \textbf{76.7} & \textbf{97.2} & \textbf{62.8} & \textbf{60.2} & \textbf{74.2}\\
      \hline

    \end{tabular}
  }
\end{table}

Experimental results consistently show that our proposed Multi-round Thinking method effectively enhances reasoning performance across diverse benchmarks. As detailed in Table 1, each evaluated model showed notable improvement when transitioning from Round 1 to Round 2 reasoning.

Specifically, for the Deepseek-R1 model, accuracy improved from 79.7\% to 82.0\% on the AIME 2024 benchmark, remained consistently high at 97.6\% on MATH-500, increased from 74.0\% to 74.8\% on GPQA-Diamond, and improved from 65.3\% to 67.1\% on LiveCodeBench.

For the QwQ-32B model \citep{qwq32b}, notable gains were achieved, with accuracy rising from 80.3\% to 82.1\% on AIME 2024 \citep{maa_aime_2024}, 97.2\% to 97.8\% on MATH-500, 63.0\% to 64.7\% on GPQA-Diamond, and 65.9\% to 67.2\% on LiveCodeBench.

Further, we evaluated our self-trained AM-Distill-Qwen-32B model, a 32B model built upon the Qwen2.5-32B \citep{qwen2.5} architecture and trained using distilled data from the DeepSeek-R1 model (refer to \citep{AM-DeepSeek-R1-Distilled-1.4M} for distillation details). Experimental results demonstrated robust performance improvements with Multi-round Thinking: accuracy increased from 72.8\% to 76.7\% on AIME 2024, from 96.2\% to 97.2\% on MATH-500, from 62.3\% to 62.8\% on GPQA-Diamond, and from 58.3\% to 60.2\% on LiveCodeBench.

Building upon this, we further examine the performance trajectory of QwQ-32B across four rounds of iterative thinking, as visualized in Figure~\ref{fig:QwQ-32B res}. The model exhibits a clear and steady upward trend across all benchmarks.

From Round 1 to Round 4, QwQ-32B's performance on AIME 2024 improves from 80.3\% to 83.1\%, indicating enhanced capability in competition-level mathematical reasoning. On the MATH-500 dataset, performance remains consistently high, fluctuating slightly within the 97.2\%–97.8\% range.

Substantial gains are observed on reasoning-heavy benchmarks like GPQA-Diamond, where accuracy increases from 65.9\% to 68.1\% over four rounds. Similarly, LiveCodeBench scores rise from 63.0\% to 65.9\%, reflecting a notable enhancement in code understanding and generation tasks.

These empirical results highlight the consistent advantage provided by the Multi-round Thinking methodology, underscoring its efficacy in iteratively refining reasoning processes, correcting earlier mistakes, and substantially boosting model performance across challenging reasoning tasks.

In summary, our results strongly indicate that Multi-round Thinking consistently improves the reasoning performance of LLMs across various tasks. Particularly notable is its effectiveness on tasks demanding complex, iterative reasoning such as mathematics competitions and coding benchmarks. Moreover, the incremental and sustained improvements observed over multiple rounds underscore the robustness of this simple yet effective test-time scaling strategy. These findings suggest that Multi-round Thinking offers an efficient pathway to enhance model accuracy without additional training overhead, thus highlighting its practical value for real-world deployment and opening promising avenues for future research in test-time scaling methods.

\subsubsection{Analysis of Word Frequency Changes}
To better understand how the model’s reasoning behavior evolves through multi-round thinking, we conduct a lexical analysis focusing on four discourse markers: but, wait, maybe, and therefore. These words serve as linguistic signals for hesitation (but, wait, maybe) or decisiveness (therefore), and tracking their usage reveals insights into the model’s confidence and reasoning dynamics.

\begin{figure}[h!]
    \centering
    \includegraphics[width=0.6\linewidth]{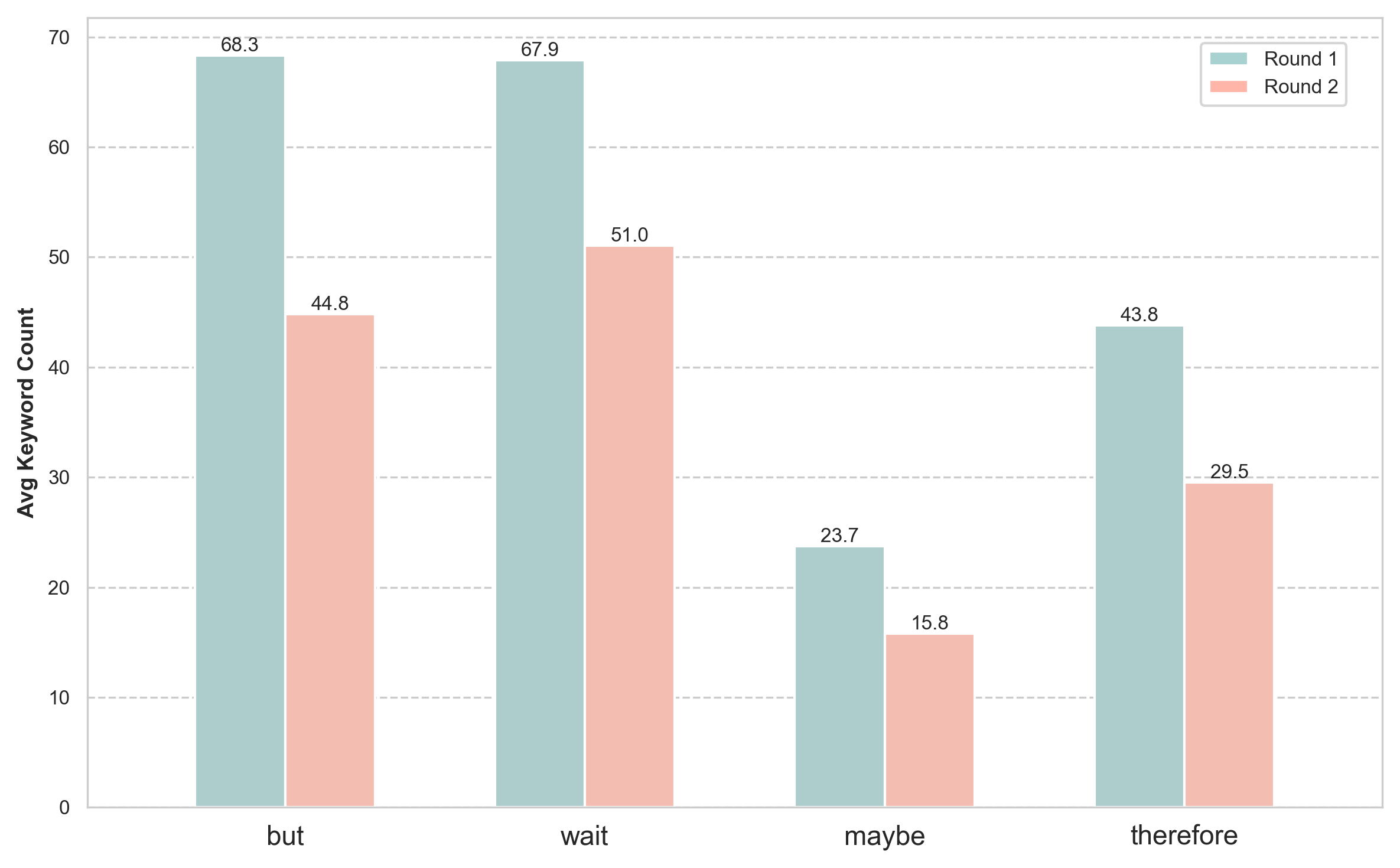}
    \caption{Overall change in word frequency across all AIME 2024 examples.}
    \label{fig:Overall change in word frequency in Multi-round Thinking}
\end{figure}

\begin{figure}[h!]
    \centering
    \includegraphics[width=0.9\linewidth]{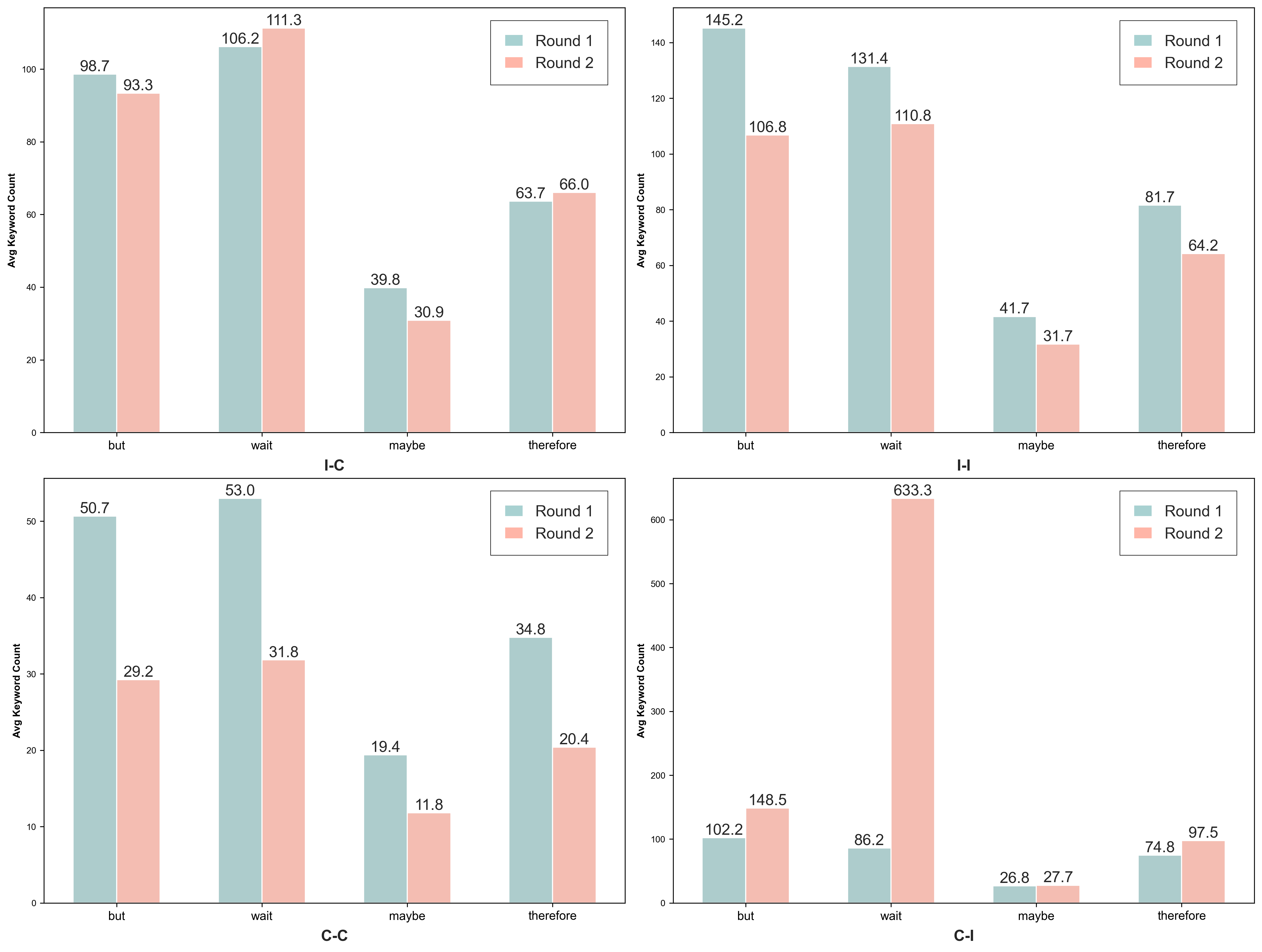}
    \caption{Changes in average word frequency across different reasoning trajectories. Each subplot shows the average frequency of four indicative words — \textit{but}, \textit{wait}, \textit{maybe}, and \textit{therefore} — in Round 1 vs. Round 2, grouped by response type: 
    I-C (Incorrect → Correct), I-I (Incorrect → Incorrect), C-C (Correct → Correct), and C-I (Correct → Incorrect). 
}
    \label{fig:Word frequency in Multi-round Thinking}
\end{figure}

Figure~\ref{fig:Overall change in word frequency in Multi-round Thinking} presents the overall average usage frequency of these keywords across all AIME 2024 test samples. From Round 1 to Round 2, we observe consistent declines in the frequency of hesitation-related words. Specifically, but decreases from 68.3 to 44.8, wait from 67.9 to 51.0, and maybe from 23.7 to 15.8. Even therefore, although a more conclusive word, experiences a drop from 43.8 to 29.5, but the relative reduction is smaller. This suggests that, overall, the model adopts more concise and assertive phrasing in Round 2 responses.

To analyze this shift in greater detail, Figure~\ref{fig:Word frequency in Multi-round Thinking} breaks down average keyword usage by answer trajectory: Incorrect→Correct (I-C), Incorrect→Incorrect (I-I), Correct→Correct (C-C), and Correct→Incorrect (C-I). In most groups, there is a significant drop in but, wait, and maybe from Round 1 to Round 2, reinforcing the observation that models tend to suppress uncertain or self-interruptive phrasing after one round of reflection. For example, in the I-I group, but drops from 145.2 to 106.8 and wait from 131.4 to 110.8, indicating that even when the model fails in both rounds, it still shifts toward more direct expression.

Interestingly, in the I-C group where the model corrects its earlier error, we observe an increase in the use of wait and therefore. This suggests a more thoughtful and deliberate step-by-step reanalysis process. The rise in therefore (from 63.7 to 66.0) also reflects the model’s increased confidence in arriving at the correct conclusion.

Together, these patterns suggest that multi-round thinking helps the model become more confident, fluent, and decisive in its responses—reducing hedging and strengthening clarity.

\subsubsection{Analysis of Response Length}

\begin{table}[htbp]
  \caption{Analysis of response length across multiple reasoning rounds for QwQ-32B, demonstrating decreasing token lengths as the model iteratively refines answers.}
  \vspace{0.5em}
  \centering
  \renewcommand{\arraystretch}{1.2}
  \label{tab:model_performance_len}
  \resizebox{\textwidth}{!}{
    \begin{tabular}{l|>{\centering\arraybackslash}m{\dimexpr0.07\textwidth\relax}||
                      M{\dimexpr0.12\textwidth\relax}
                      M{\dimexpr0.12\textwidth\relax}
                      M{\dimexpr0.12\textwidth\relax}
                      M{\dimexpr0.18\textwidth\relax}|
                      M{\dimexpr0.12\textwidth\relax}}
      \hline
      \multicolumn{1}{l|}{\textbf{Model}} & \textbf{Round} & \textbf{AIME 2024} & \textbf{MATH-500} & \textbf{GPQA-Diamond} & \textbf{LiveCodeBench} & \textbf{Average}\\[1.5ex]
      \hline
      \multirow{2}{*}{\textbf{QwQ-32B}} 
        & 1 & 13566.1 & 8489.9 & 13860.5 & 4473.0 & 10097.4\\
        & 2 & 9607.9 & 5540.4 & 11043.9 & 3200.9 & 7348.3\\
        & 3 & 8630.0 & 5287.7 & 10368.0 & 3012.7 & 6824.6\\
        & 4 & 8654.8 & 4948.0 & 9674.5 & 2920.8 & 6549.5\\
      \hline

    \end{tabular}
  }
\end{table}

We analyzed the generation lengths of the model across different inference rounds, as shown in Table \ref{tab:model_performance_len}.

As the number of inference rounds increases, the generation length tends to decrease. Moreover, there is a correlation between performance improvement and the reduction in generation length—the greater the performance gain, the more significant the reduction. For example, from Round 1 to Round 2, the average score improves by 1.4 points while the average generation length decreases by 2749.1 tokens; from Round 2 to Round 3, there is minimal improvement in performance, while the average generation length decreases by only 675.9 tokens.

\begin{figure*}[hbt]
\vspace{0.1cm}
\centering
\setlength{\abovecaptionskip}{0.1cm} 
\includegraphics[scale=0.45]{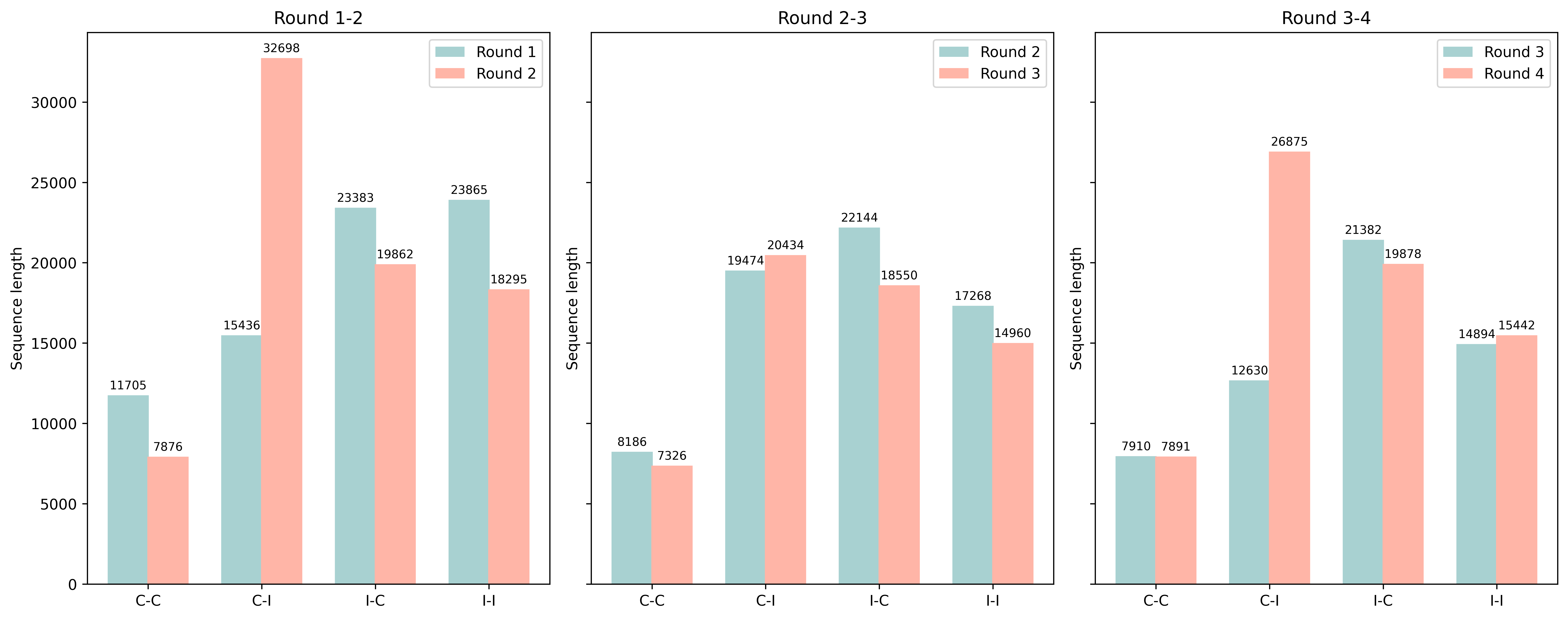}
\caption{Changes in response length across reasoning rounds on the AIME 2024 dataset (QwQ-32B model). Labels represent the correctness trajectory from Round 1 to Round 2: ``C'' = Correct, ``I'' = Incorrect. For example, ``C–I'' indicates responses that were correct initially but became incorrect in the next round.}
\label{answer_length_variation}
\end{figure*}


As shown in Figure \ref{answer_length_variation}, we analyzed the changes in response lengths for the same questions across different inference rounds. We found that when the model answered correctly in the previous round but incorrectly in the current round, the inference length increased significantly. This trend is consistent with the word frequency analysis in Figure \ref{fig:Overall change in word frequency in Multi-round Thinking}, where the frequency of the word "wait" rises notably, suggesting increased uncertainty in the model’s reasoning. In contrast, when the model answers correctly in both the previous and current rounds, it becomes more confident, leading to a substantial reduction in inference length.


\subsection{A Preliminary Experiment with Supervised Fine-tuning (SFT)}


To further enhance the robustness of Multi-round Thinking, we explored combining it with supervised fine-tuning (SFT). The key idea was to reduce error propagation from earlier reasoning rounds by explicitly training the model to rectify previously incorrect answers.

Specifically, our supervised fine-tuning process involved the following steps:

\begin{itemize}
\item \textbf{Data selection}: We selected challenging mathematical and programming tasks from open datasets, ensuring each could be independently verified.
\item \textbf{Data generation}: We employed the DeepSeek-R1 model iteratively on these tasks using Multi-round Thinking until a correct answer was verified, forming a dataset of approximately 100,000 examples.
\item \textbf{Model training}: This dataset was used to fine-tune our AM-32B model in an initial round of supervised training.
\end{itemize}

\begin{table}[htbp]
  \caption{Model Performance with Average(pass@1)}
  \vspace{0.5em}
  \centering
  \renewcommand{\arraystretch}{1.2}
  \label{tab:model_performance}
  \resizebox{\textwidth}{!}{
    \begin{tabular}{l|>{\centering\arraybackslash}m{\dimexpr0.07\textwidth\relax}||
                      M{\dimexpr0.12\textwidth\relax}
                      M{\dimexpr0.12\textwidth\relax}
                      M{\dimexpr0.12\textwidth\relax}
                      M{\dimexpr0.18\textwidth\relax}|
                      M{\dimexpr0.12\textwidth\relax}}
      \hline
      \multicolumn{1}{l|}{\textbf{Model}} & \textbf{Round} & \textbf{AIME 2024} & \textbf{MATH-500} & \textbf{GPQA-Diamond} & \textbf{LiveCodeBench} & \textbf{Average}\\[1.5ex]
      \hline
      \multirow{2}{*}{\textbf{AM-32B}} 
        & 1 & 72.8 & 96.2 & 62.3 & 58.3 & 72.4\\
        & \textbf{2} & \textbf{76.7} & \textbf{97.2} & \textbf{62.8} & \textbf{60.2} & \textbf{74.2}\\
        & 2(SFT) & 75.9 & 97.0 & 63.0 & 57.9 & 73.5\\
      \hline
    \end{tabular}
  }
\end{table}

After SFT, we conducted further experiments based on the AM-32B Round2 reasoning outputs. While this preliminary fine-tuning did not lead to performance improvements in our current evaluation (see Table 3), it opens up promising directions for future research in leveraging high-quality reasoning data to enhance Multi-round Thinking.

\section{Discussion and Conclusion}
In this study, we proposed Multi-round Thinking, a straightforward yet effective test-time scaling strategy designed to enhance the reasoning capabilities of large language models (LLMs). Inspired by human cognitive processes, this iterative approach allows models to refine their reasoning by independently reconsidering their previous answers, significantly mitigating cognitive inertia and correcting initial reasoning errors. Our extensive experiments demonstrated consistent and substantial improvements across challenging benchmarks, including AIME 2024, GPQA-Diamond, MATH-500, and LiveCodeBench. For instance, accuracy improved by more than 2 percentage points on complex mathematical competition tasks, underscoring the broad applicability and practical value of this approach.

Further analysis revealed that multi-round reasoning not only improved accuracy but also made the models' reasoning more concise and confident. Specifically, we observed a reduction in uncertainty markers (such as "but", "wait", and "maybe") and shorter responses, reflecting increased model clarity and decisiveness in reasoning. These linguistic insights indicate that iterative thinking aligns closely with human cognitive patterns, enhancing the transparency and interpretability of LLM behaviors.

While preliminary experiments integrating supervised fine-tuning (SFT) did not immediately yield additional improvements, they highlighted crucial considerations for future research—particularly regarding the quality of training data and fine-tuning strategies tailored explicitly for iterative reasoning. Exploring these directions further promises significant theoretical and practical benefits, potentially unlocking even greater reasoning capabilities in LLMs.

In practical product applications, adopting a ``think twice'' approach can conveniently incorporate the first-round response as part of the thinking process itself, effectively realizing performance gains. However, this inevitably introduces additional waiting time during the thinking phase.

In summary, Multi-round Thinking represents a practical, efficient, and universally applicable method for improving LLM reasoning without additional training overhead. This research opens valuable pathways for future exploration and offers immediate utility for both academia and industry in the ongoing quest for more robust, reliable, and explainable AI reasoning.

\bibliographystyle{plainnat}
\bibliography{references}

\begin{thebibliography}{26}
\providecommand{\natexlab}[1]{#1}
\providecommand{\url}[1]{\texttt{#1}}
\expandafter\ifx\csname urlstyle\endcsname\relax
  \providecommand{\doi}[1]{doi: #1}\else
  \providecommand{\doi}{doi: \begingroup \urlstyle{rm}\Url}\fi

\bibitem[Amodei et~al.(2016)Amodei, Olah, Steinhardt, Christiano, Schulman, and Mané]{amodei2016concreteproblemsaisafety}
Dario Amodei, Chris Olah, Jacob Steinhardt, Paul Christiano, John Schulman, and Dan Mané.
\newblock Concrete problems in ai safety, 2016.
\newblock URL \url{https://arxiv.org/abs/1606.06565}.

\bibitem[Chen et~al.(2025)Chen, Qin, Liu, Peng, Guan, Wang, Hu, Zhou, Gao, and Che]{chen2025reasoningerasurveylong}
Qiguang Chen, Libo Qin, Jinhao Liu, Dengyun Peng, Jiannan Guan, Peng Wang, Mengkang Hu, Yuhang Zhou, Te~Gao, and Wanxiang Che.
\newblock Towards reasoning era: A survey of long chain-of-thought for reasoning large language models, 2025.
\newblock URL \url{https://arxiv.org/abs/2503.09567}.

\bibitem[Choi et~al.(2023)Choi, Fang, Wang, and Song]{choi2023kctsknowledgeconstrainedtreesearch}
Sehyun Choi, Tianqing Fang, Zhaowei Wang, and Yangqiu Song.
\newblock Kcts: Knowledge-constrained tree search decoding with token-level hallucination detection, 2023.
\newblock URL \url{https://arxiv.org/abs/2310.09044}.

\bibitem[DeepSeek-AI(2025)]{deepseekai2025deepseekr1incentivizingreasoningcapability}
DeepSeek-AI.
\newblock Deepseek-r1: Incentivizing reasoning capability in llms via reinforcement learning, 2025.
\newblock URL \url{https://arxiv.org/abs/2501.12948}.

\bibitem[Diao et~al.(2024)Diao, Wang, Lin, Pan, Liu, and Zhang]{diao2024activepromptingchainofthoughtlarge}
Shizhe Diao, Pengcheng Wang, Yong Lin, Rui Pan, Xiang Liu, and Tong Zhang.
\newblock Active prompting with chain-of-thought for large language models, 2024.
\newblock URL \url{https://arxiv.org/abs/2302.12246}.

\bibitem[Hoffmann et~al.(2022)Hoffmann, Borgeaud, Mensch, Buchatskaya, Cai, Rutherford, de~Las~Casas, Hendricks, Welbl, Clark, Hennigan, Noland, Millican, van~den Driessche, Damoc, Guy, Osindero, Simonyan, Elsen, Rae, Vinyals, and Sifre]{hoffmann2022trainingcomputeoptimallargelanguage}
Jordan Hoffmann, Sebastian Borgeaud, Arthur Mensch, Elena Buchatskaya, Trevor Cai, Eliza Rutherford, Diego de~Las~Casas, Lisa~Anne Hendricks, Johannes Welbl, Aidan Clark, Tom Hennigan, Eric Noland, Katie Millican, George van~den Driessche, Bogdan Damoc, Aurelia Guy, Simon Osindero, Karen Simonyan, Erich Elsen, Jack~W. Rae, Oriol Vinyals, and Laurent Sifre.
\newblock Training compute-optimal large language models, 2022.
\newblock URL \url{https://arxiv.org/abs/2203.15556}.

\bibitem[Jain et~al.(2024)Jain, Han, Gu, Li, Yan, Zhang, Wang, Solar-Lezama, Sen, and Stoica]{jain2024livecodebench}
Naman Jain, King Han, Alex Gu, Wen-Ding Li, Fanjia Yan, Tianjun Zhang, Sida Wang, Armando Solar-Lezama, Koushik Sen, and Ion Stoica.
\newblock Livecodebench: Holistic and contamination free evaluation of large language models for code.
\newblock \emph{arXiv preprint arXiv:2403.07974}, 2024.

\bibitem[Kaplan et~al.(2020)Kaplan, McCandlish, Henighan, Brown, Chess, Child, Gray, Radford, Wu, and Amodei]{kaplan2020scalinglawsneurallanguage}
Jared Kaplan, Sam McCandlish, Tom Henighan, Tom~B. Brown, Benjamin Chess, Rewon Child, Scott Gray, Alec Radford, Jeffrey Wu, and Dario Amodei.
\newblock Scaling laws for neural language models, 2020.
\newblock URL \url{https://arxiv.org/abs/2001.08361}.

\bibitem[Langosco et~al.(2023)Langosco, Koch, Sharkey, Pfau, Orseau, and Krueger]{langosco2023goalmisgeneralizationdeepreinforcement}
Lauro Langosco, Jack Koch, Lee Sharkey, Jacob Pfau, Laurent Orseau, and David Krueger.
\newblock Goal misgeneralization in deep reinforcement learning, 2023.
\newblock URL \url{https://arxiv.org/abs/2105.14111}.

\bibitem[Levi(2024)]{levi2024simplemodelinferencescaling}
Noam Levi.
\newblock A simple model of inference scaling laws, 2024.
\newblock URL \url{https://arxiv.org/abs/2410.16377}.

\bibitem[Li et~al.(2025)Li, Dong, Jin, Zhang, Zhou, Zhu, Zhang, and Dou]{li2025searcho1agenticsearchenhancedlarge}
Xiaoxi Li, Guanting Dong, Jiajie Jin, Yuyao Zhang, Yujia Zhou, Yutao Zhu, Peitian Zhang, and Zhicheng Dou.
\newblock Search-o1: Agentic search-enhanced large reasoning models, 2025.
\newblock URL \url{https://arxiv.org/abs/2501.05366}.

\bibitem[Lightman et~al.(2023)Lightman, Kosaraju, Burda, Edwards, Baker, Lee, Leike, Schulman, Sutskever, and Cobbe]{lightman2023letsverifystepstep}
Hunter Lightman, Vineet Kosaraju, Yura Burda, Harri Edwards, Bowen Baker, Teddy Lee, Jan Leike, John Schulman, Ilya Sutskever, and Karl Cobbe.
\newblock Let's verify step by step, 2023.
\newblock URL \url{https://arxiv.org/abs/2305.20050}.

\bibitem[MAA(2024)]{maa_aime_2024}
MAA.
\newblock American invitational mathematics examination - aime.
\newblock \url{https://maa.org/math-competitions/american-invitational-mathematics-examination-aime}, feb 2024.
\newblock Accessed in February 2024, from American Invitational Mathematics Examination - AIME 2024.

\bibitem[Muennighoff et~al.(2025)Muennighoff, Yang, Shi, Li, Fei-Fei, Hajishirzi, Zettlemoyer, Liang, Candès, and Hashimoto]{muennighoff2025s1simpletesttimescaling}
Niklas Muennighoff, Zitong Yang, Weijia Shi, Xiang~Lisa Li, Li~Fei-Fei, Hannaneh Hajishirzi, Luke Zettlemoyer, Percy Liang, Emmanuel Candès, and Tatsunori Hashimoto.
\newblock s1: Simple test-time scaling, 2025.
\newblock URL \url{https://arxiv.org/abs/2501.19393}.

\bibitem[OpenAI(2024{\natexlab{a}})]{OpenAI2024}
OpenAI.
\newblock Learning to reason with llms, 2024{\natexlab{a}}.
\newblock URL \url{https://openai.com/index/learning-to-reason-with-llms/}.

\bibitem[OpenAI(2024{\natexlab{b}})]{openai2024openaio1card}
OpenAI.
\newblock Openai o1 system card, 2024{\natexlab{b}}.
\newblock URL \url{https://arxiv.org/abs/2412.16720}.

\bibitem[Qin et~al.(2024)Qin, Li, Zou, Liu, Xia, Huang, Ye, Yuan, Liu, Li, and Liu]{qin2024o1replicationjourneystrategic}
Yiwei Qin, Xuefeng Li, Haoyang Zou, Yixiu Liu, Shijie Xia, Zhen Huang, Yixin Ye, Weizhe Yuan, Hector Liu, Yuanzhi Li, and Pengfei Liu.
\newblock O1 replication journey: A strategic progress report -- part 1, 2024.
\newblock URL \url{https://arxiv.org/abs/2410.18982}.

\bibitem[Qwen(2024)]{qwen2.5}
Qwen.
\newblock Team qwen2.5: A party of foundation models, September 2024.
\newblock URL \url{https://qwenlm.github.io/blog/qwen2.5/}.

\bibitem[Rein et~al.(2023)Rein, Hou, Stickland, Petty, Pang, Dirani, Michael, and Bowman]{rein2023gpqagraduatelevelgoogleproofqa}
David Rein, Betty~Li Hou, Asa~Cooper Stickland, Jackson Petty, Richard~Yuanzhe Pang, Julien Dirani, Julian Michael, and Samuel~R. Bowman.
\newblock Gpqa: A graduate-level google-proof q\&a benchmark, 2023.
\newblock URL \url{https://arxiv.org/abs/2311.12022}.

\bibitem[Team(2025)]{qwq32b}
Qwen Team.
\newblock Qwq-32b: Embracing the power of reinforcement learning, March 2025.
\newblock URL \url{https://qwenlm.github.io/blog/qwq-32b/}.

\bibitem[Wang et~al.(2024)Wang, Li, Shao, Xu, Dai, Li, Chen, Wu, and Sui]{wang2024mathshepherdverifyreinforcellms}
Peiyi Wang, Lei Li, Zhihong Shao, R.~X. Xu, Damai Dai, Yifei Li, Deli Chen, Y.~Wu, and Zhifang Sui.
\newblock Math-shepherd: Verify and reinforce llms step-by-step without human annotations, 2024.
\newblock URL \url{https://arxiv.org/abs/2312.08935}.

\bibitem[Wei et~al.(2023)Wei, Wang, Schuurmans, Bosma, Ichter, Xia, Chi, Le, and Zhou]{wei2023chainofthoughtpromptingelicitsreasoning}
Jason Wei, Xuezhi Wang, Dale Schuurmans, Maarten Bosma, Brian Ichter, Fei Xia, Ed~Chi, Quoc Le, and Denny Zhou.
\newblock Chain-of-thought prompting elicits reasoning in large language models, 2023.
\newblock URL \url{https://arxiv.org/abs/2201.11903}.

\bibitem[Wu et~al.(2025)Wu, Sun, Li, Welleck, and Yang]{wu2025inferencescalinglawsempirical}
Yangzhen Wu, Zhiqing Sun, Shanda Li, Sean Welleck, and Yiming Yang.
\newblock Inference scaling laws: An empirical analysis of compute-optimal inference for problem-solving with language models, 2025.
\newblock URL \url{https://arxiv.org/abs/2408.00724}.

\bibitem[Yang et~al.(2025)Yang, Ma, Lin, and Wei]{yang2025thinkingoptimalscalingtesttimecompute}
Wenkai Yang, Shuming Ma, Yankai Lin, and Furu Wei.
\newblock Towards thinking-optimal scaling of test-time compute for llm reasoning, 2025.
\newblock URL \url{https://arxiv.org/abs/2502.18080}.

\bibitem[Zhao et~al.(2025)Zhao, Wang, Peng, Zhao, Tian, Chen, Ji, and Li]{AM-DeepSeek-R1-Distilled-1.4M}
Han Zhao, Haotian Wang, Yiping Peng, Sitong Zhao, Xiaoyu Tian, Shuaiting Chen, Yunjie Ji, and Xiangang Li.
\newblock 1.4 million open-source distilled reasoning dataset to empower large language model traning, 2025.
\newblock URL \url{https://github.com/a-m-team/a-m-models/blob/main/docs/AM-DeepSeek-R1-Distilled-Dataset.pdf}.

\bibitem[Zhou et~al.(2024)Zhou, Yan, Shlapentokh-Rothman, Wang, and Wang]{zhou2024languageagenttreesearch}
Andy Zhou, Kai Yan, Michal Shlapentokh-Rothman, Haohan Wang, and Yu-Xiong Wang.
\newblock Language agent tree search unifies reasoning acting and planning in language models, 2024.
\newblock URL \url{https://arxiv.org/abs/2310.04406}.

\end{thebibliography}

\clearpage
\appendix

\section{Example}
\label{a}
\begin{figure}[h!]
    \centering
    \includegraphics[width=0.9\linewidth]{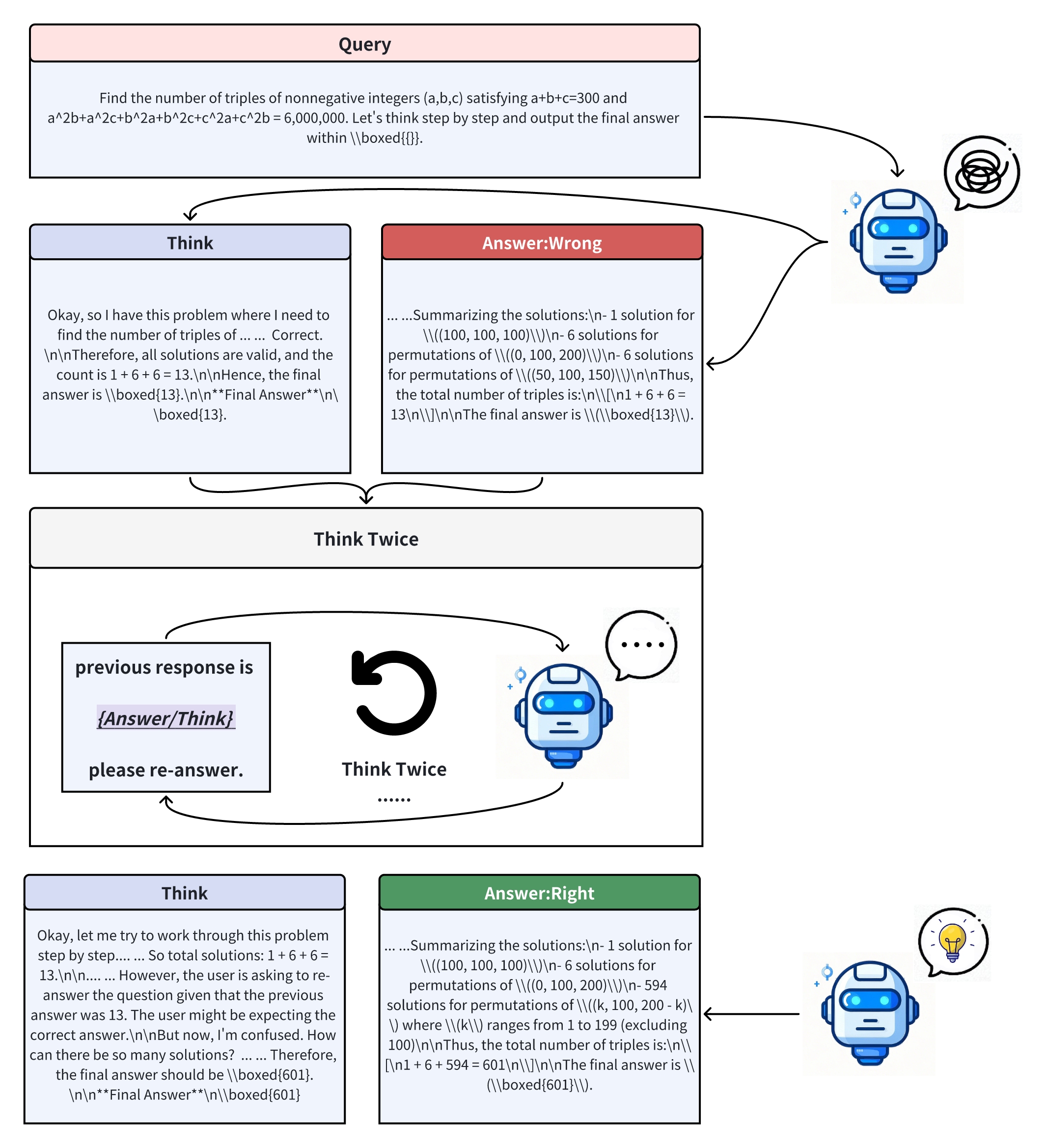}
    \caption{Illustration of the ``Think Twice'' Strategy in Multi-round Reasoning.}
    \label{fig:Think Twice example}
\end{figure}

The model first provides an incorrect answer by following its initial reasoning chain.Upon invoking the \texttt{Think Twice} mechanism, it is explicitly prompted to reassess its prior response. The model then revisits its reasoning, identifies the over-simplified solution space, and produces a corrected answer with a significantly expanded and accurate enumeration. This process highlights the effectiveness of forcing self-reflection to catch subtle counting mistakes.

\end{document}